# Seuillage par hystérésis pour le test de photo-consistance des voxels dans le cadre de la reconstruction 3D

Mohamed Chafik Bakkay, Walid Barhoumi et Ezzeddine Zagrouba

Equipe de Recherche Systèmes Intelligents en Imagerie et Vision Artificielle (SIIVA), Laboratoire RIADI
Institut Supérieur d'Informatique, 2 Rue Abou Rayhane Bayrouni, 2080 Ariana, Tunisie
`med.chafik@hotmail.fr, walid.barhoumi@laposte.net, ezzeddine.zagrouba@fsm.rnu.tn`

## RÉSUMÉ

La coloration des voxels est une méthode populaire de reconstruction d'un modèle à trois dimensions d'une surface volumétrique à partir d'un ensemble d'images 2D calibrées. Cependant, la qualité de reconstruction de l'algorithme dépend fortement d'une étape de seuillage. Pourtant, cette méthode est largement utilisée en raison de sa simplicité et son coût de calcul faible. Nous sommes retournés à cette méthode afin d'y proposer une amélioration dans l'étape de seuillage qui sera entièrement automatisée. En effet, l'information géométrique est implicitement intégrée en utilisant un seuillage par hystérésis qui prend en considération la connexité des voxels colorés. En plus, l'ambiguïté dans le choix des seuils est minimisée grâce à l'utilisation d'un degré d'appartenance flou de chaque voxel à la classe des voxels photo-consistants, et à la détermination automatique des seuils en fonction du nombre des images sur lesquelles le voxel est projeté.

## Mots-clés

coloration des voxels, reconstruction 3D, seuillage par hystérésis, suites adjacentes, logique floue.

## 1. INTRODUCTION

La reconstruction d'une scène volumétrique à partir d'un ensemble d'images est un problème très intéressant et ouvert dans le domaine de la vision par ordinateur. Elle peut avoir plusieurs applications dans des domaines différents tels que la robotique, l'infographie et la réalité virtuelle. Par exemple, il y a un besoin croissant pour la reconstruction des modèles 3D dans le domaine des jeux vidéo et dans le domaine de l'animation qui ont eu un développement significatif pendant la dernière décennie. Ainsi, de nombreuses approches ont été proposées pour résoudre ce problème. La coloration des voxels, introduite par Seitz et Dyer [1], est l'une des approches de reconstruction 3D qui déduisent la forme de l'objet à partir de la photo-consistance. C'est une technique populaire pour la reconstruction 3D dont l'idée de base est de classer chaque élément 3D (voxel), appartenant à un volume discrétisé englobant la scène filmée, afin de décider s'il appartient à la surface de l'objet 3D filmé ou non. Ceci renvient à tester la photo-consistance de chaque voxel, qui est la plus ancienne et la plus simple propriété photométrique d'une scène dans la littérature de la stéréovision. Elle prétend que, dans le cas de l'appartenance du voxel au volume 3D filmé, les pixels correspondants sont supposés avoir la même couleur dans les projections 2D relatives. En d'autres termes, un point sur une surface doit être vu avec des couleurs similaires quand il n'est pas occlut. Cette propriété est basée sur un couple d'hypothèses. D'une part, les objets de la scène devraient avoir des surfaces Lambertiennes. D'autre part, la projection d'un point 3D quelconque de la scène sur les images correspondantes peut être efficacement calculée [2]. Pour ce faire, l'algorithme divise l'espace de la scène 3D en une grille de voxels. Puis, il traverse chaque voxel pour tester sa photo-consistance par laquelle le voxel est choisi soit pour être coloré, soit pour être éliminé. Pour gérer l'occlusion correctement, l'algorithme traverse l'espace des voxels dans un ordre adéquat. En effet, les caméras utilisées sont placées sur le même côté d'un plan, et les voxels les plus proches de la caméra sont visités en premier. Cela garantit que le voxel ne peut pas être occlut par un voxel non visité. La reconstruction résultante est appelée le photo-hull [3], qui est défini comme étant le volume maximal qui est photo-consistant avec l'ensemble de vues d'entrée. En particulier, afin d'estimer la photo-consistance d'un voxel donné dans l'espace 3D, il est projeté sur les images d'entrée pour récupérer les valeurs de couleur dans la région où le voxel a été projeté. Sous l'hypothèse de réflectance Lambertienne, un voxel ayant presque la même couleur dans toutes les images dans lesquelles il est visible (voxel consistant), devrait avoir cette couleur. Sinon, le voxel doit être enlevé.

La plupart des techniques de coloration des voxels souffrent de plusieurs limitations importantes, notamment les problèmes liés aux tests de photo-consistance. Tout d'abord, elles supposent une texture suffisante et sont généralement très sensibles aux erreurs de bruit et de calibration. En outre, la représentation voxel ignore la continuité de la forme, ce qui rend la mise en vigueur de la souplesse et la cohérence spatiale très difficile. Un autre inconvénient des tests existants de photo-consistance est que la qualité de la reconstruction est très sensible à la valeur des seuils utilisés. Par ailleurs, les voxels éliminés ne peuvent pas être récupérés et peuvent être éliminés par erreur dans un effet de cascade. Cette limitation est partiellement atténuée par la méthode de sculpture d'espace probabiliste [4]. En outre, le choix du seuil global pour la variance de couleur est souvent problématique. Pour cela, il y a eu quelques tentatives pour alléger ces contraintes [5] [6]. Dans ce cadre, nous proposons dans ce papier un nouveau test de photo-consistance qui minimise l'influence des seuils, et ceci en utilisant la logique floue et un seuillage automatique adaptatif, tout en intégrant l'information géométrique par l'utilisation d'un seuillage par hystérésis.

Le reste de ce papier est structuré comme suit. Dans la section 2, nous présentons un aperçu de la technique proposée tout en résumant ses principales contributions. Les résultats expérimentaux sont présentés dans la section 3 pour démontrer l'efficacité de la solution proposée. La section 4 conclut le papier et présente quelques directions pour les futurs travaux.

## 2. TECHNIQUE PROPOSÉE

Ayant les transformations 3D fournies pour plusieurs caméras calibrées, chaque voxel $V_i$ est projeté sur les images d'entrée ($I_1...I_n$). Les Voxels sont parcourus dans un ordre qui satisfait la contrainte de visibilité ordinale. Puis, tous les pixels ($p_1...p_k$) qui n'ont pas encore été marqués sont collectés. S'il n'y a pas de pixels, le voxel $V_i$ est éliminé. Le nombre N des images dans lesquelles $V_i$ est visible est aussi défini. Après le tri de toutes les valeurs des pixels afin d'éliminer les valeurs aberrantes, le degré d'appartenance de $V_i$, à la classe des voxels photo-consistants, est défini par une fonction d'évaluation de l'homogénéité. Ensuite, une estimation automatique des seuils, en fonction du nombre N, est utilisée pour définir automatiquement le seuil supérieur et le seuil inférieur, respectivement $T_{haut}$ et $T_{bas}$, qui seront utilisés lors du seuillage par hystérésis. En effet, le seuillage est plus stricte lorsque N augmente. Enfin, le seuillage par hystérésis est utilisé pour déterminer si le voxel est photo-consistant ou non. Si le voxel $V_i$ est photo-consistant, alors il est accepté et la couleur moyenne de l'ensemble de pixels ($p_1...p_k$) est lui affectée. Sinon, le voxel $V_i$ est éliminé. Les pixels sont ensuite marqués pour indiquer qu'ils sont déjà utilisés.

### 2.1 Projection d'un voxel

Tout voxel $V_i$ est projeté sur chaque image I comme suit. Tout d'abord, les huit sommets du voxel sont projetés à partir des coordonnées du monde réel vers les coordonnées de l'image à l'aide du modèle de la caméra [15, 16]. Ensuite, la boîte englobante autour de ces projections est utilisée pour approcher la forme réelle du voxel. Les pixels ($p_1...p_k$) qui sont dans cette boîte englobante et qui n'ont pas encore été marqués sont collectés. Le nombre N des images dans lesquelles $V_i$ est visible, est aussi déterminé. Cette valeur permettra de définir automatiquement et d'une manière adaptative les seuils pour le test de photo-consistance.

### 2.2 Évaluation de l'homogénéité

Le degré d'appartenance $\mu(V_i)$ d'un voxel $V_i$, à la classe des voxels photo-consistants, est défini par une fonction d'évaluation de l'homogénéité basée sur la logique floue. Pour ce faire, l'ensemble non vide de couleurs, ($p_1...p_k$), est out d'abord extrait à partir des images sur lesquelles le voxel $V_i$ est projeté. Ensuite, cet ensemble est trié selon les valeurs de pixels afin d'éliminer les valeurs aberrantes. Le degré d'appartenance est enfin calculé comme le quotient entre la valeur minimale et celle maximale parmi toutes les valeurs de pixels retenus.

### 2.3 Éstimation automatique des seuils

Le seuillage par hystérésis devrait être plus stricte lorsque N augmente. Ainsi, lorsque N augmente, $T_{haut}$ devrait diminuer et $T_{bas}$ devrait augmenter. Pour cela, nous avons modélisé les deux seuils $T_{haut}$ et $T_{bas}$ par deux suites adjacentes qui dépendent de N. En effet, deux suites adjacentes sont telles que l'une est croissante et l'autre est décroissante et les termes des deux se rapprochent lorsque N tend vers l'infini. Cela signifie qu'elles sont convergentes et qu'elles convergent vers la même limite. En effet, nous supposons que $T_{bas} = S_{2n}$ et $T_{haut} = S_{2n+1}$ avec $\{S_n\}$ est une suite harmonique alternée. Puisque $\{S_{2n}\}$ est croissante et $\{S_{2n+1}\}$ est décroissante, ces deux suites sont adjacentes et elles convergent vers la même limite qui vaut log 2. Nous avons opté pour ces suites vu que nos expérimentations nous ont prouvé qui les meilleurs résultats de reconstruction 3D, avec l'algorithme naïf (mono-seuil) de coloration des voxels, ont été enregistrés avec des valeurs de seuils aux alentours de log 2.

### 2.4 Seuillage par hystérésis

Dans cette technique de seuillage, deux seuils sont nécessaires : un seuil haut $T_{haut}$ et un seuil bas $T_{bas}$. Le seuil haut est utilisé pour sélectionner les voxels les plus photo-consistants. Ainsi, un voxel $V_i$ est dite photo-consistant si $\mu(V_i) > T_{haut}$. Un voxel est appelé non photo-consistant si $\mu(V_i) < T_{bas}$. Tous les autres voxels sont appelés des voxels candidats. L'hypothèse de la connexité des voxels photo-consistants nous permet de vérifier le voisinage d'un voxel donné et de rejeter des voxels bruyants qui ne sont pas photo-consistants. Ainsi, l'algorithme est décrit comme suit : pour chaque voxel $V_i$, éliminer ce voxel s'il n'est pas photo-consistant, le colorier s'il est photo-consistant. Sinon, si $V_i$ est un voxel candidat, déterminer son voisinage défini par 26 voisins. S'il est relié à un voxel voisin photo-consistant, colorier $V_i$; sinon, le voxel sera éliminé du model 3D envisagé.

## 3. ÉTUDE EXPÉRIMENTALE

Pour évaluer la technique proposée, nous avons utilisé les ensembles de données multi-vues de Middlebury qui ont été fournis par Seitz et al. [7] (temple et dino). Dans cette section, nous montrons quelques résultats de la technique proposée et nous présentons une évaluation objective en termes de qualité. Notre implémentation est basée sur le framework open-source de voxel coloring, qui est développé par Van de Sande [12], disponible à l'adresse http://voxelcoloring.sourceforge.net. Tout d'abord, nous avons comparé le test de photo-consistance proposé avec celui basé sur la variance [1], celui basé sur l'histogramme [18] et le test de photo-consistance adaptatif [17]. Nous présentons dans ce qui suit les résultats des ces quatre techniques sur les ensembles de données temple éparse (Figure. 1). Il est claire que la technique proposée produit la meilleure reconstruction, par rapport aux techniques comparées, pour l'ensemble de données temple éparse.

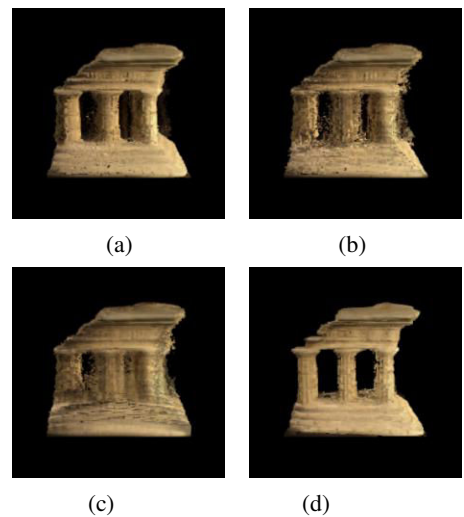

(a) (b)

(c) (d)

**Figure 1. Les résultats de la reconstruction 3D. (a) seuillage basé sur la variance (b) seuillage adaptatif (c) seuillage basé sur l'histogramme (d) technique proposée.**

En outre, la performance de la technique proposée est évaluée en utilisant la méthodologie décrite dans [7]. Cette évaluation fournit des métriques sur la précision et la complétude de la reconstruction 3D. Les reconstructions sont évaluées par comparaison géométrique avec le modèle de la vérité-terrain. Ce modèle est désigné par G et le résultat de reconstruction à évaluer est notée R. Pour mesurer la précision d'une reconstruction (à quel point R est proche de G), les distances entre les points de R et les points les plus proches sur G sont calculées. Pour mesurer la complétude (à quel point G est modélisé par R), les distances entre les points de G et les points les plus proches sur R sont calculées. Le tableau 1 résume les résultats des métriques de la précision et de la complétude de notre technique ainsi que celles d'autres techniques récentes plus complexes [15, 16]. Un seuil de précision de 90% est utilisé (i.e. une précision de 1,0 mm signifie que 90% des points sont dans 1 mm du modèle de la vérité terrain) et un seuil de 1,25 mm est utilisé pour la complétude (i.e. une complétude de 95% signifie que 95% des points sont à 1,25 mm du modèle de la vérité terrain). La précision et la complétude de notre reconstruction pour l'ensemble de données temple éparse 16-vues étaient 8.97mm et 72,1% respectivement. Les mêmes mesures pour l'ensemble de données dino éparse 16-vues étaient 8.12mm et 97,2%. En comparant la technique proposée avec d'autres techniques de reconstruction multi-vues, nous constatons que la technique proposée est entièrement automatisée, simple et a un faible coût de calcul et produit des reconstructions acceptables. Par contre, elle est moins précise que les autres techniques récentes [8, 9, 10, 11], parce que ces techniques utilisent plus de vues (16 vues) et des algorithmes plus complexes, qui possèdent des coûts de calcul élevés. En effet, la coloration des voxels ne produit pas de reconstruction complète car il y avait une contrainte sur le placement des caméras. Ainsi, nous avons utilisé seulement 8 points de vue au lieu de 16 points de vue. Si nous avions utilisé tous les points de vue, les résultats auraient changé d'une manière significative.

**Tableau 1. Les résultats de la précision et la complétude.**

|  | **Temple SparseR.** | **Dino SparseR.** |
|---|---|---|
| Notre technique | 8.97 / 72.1 % | 8.12 / 70.0 % |
| Vogiatzis [16] | 2.77 / 79.4% | 1.18 / 90.8% |
| Tran [15] | 1.53 / 85.4% | 1.26 / 89.3% |

## 4. CONCLUSION

Nous avons présenté un nouveau test automatique de photo-consistance, pour l'algorithme de coloration des voxels, qui utilise un seuillage adaptatif flou par hystérésis. Nous avons intégré l'information géométrique à l'aide du seuillage par hystérésis qui considère la connexité des voxels colorés. Nous avons aussi minimisé l'ambiguïté du choix des seuils grâce à un degré d'appartenance flou calculé pour chaque voxel, afin d'estimer la probabilité que ce voxel soit photo-consistant. En plus, la technique proposée utilise un seuillage adaptatif en fonction du nombre d'images sur lesquelles le voxel est projeté. En guise de perspectives, comme la coloration généralisée des voxels et la sculpture d'espace permettent de gérer les configurations arbitraires de la caméra pour avoir une reconstruction complète, nous allons étudier si l'application de notre technique sur ces deux méthodes fera fortifier la complétude et la précision du modèle.

## 5. RÉFÉRENCES

[1] Seitz, S. et Dyer, C. 1997. Photorealistic scene reconstruction by voxel coloring. CVPR, 1067-1073.

[2] Kutulakos, K. 2000. Approximate n-view stereo. ECCV, 67-83.

[3] Kiriakos, N., Kutulakos, K et Seitz, S. 2000. A theory of shape by space carving. IJCV, 38, 3, 198-218.

[4] Broadhurst, A., Drummond, T. et Cipolla, R. 2001. A probabilistic framework for space carving. ICCV, 388-393.

[5] Treuille, A., Hertzmann, A. et Seitz, S. 2004. Example-based stereo with general BRDFs. ECCV. 2, 457-469.

[6] Yang, R., Pollefeys, M. et Welch, G. 2003. Dealing with textureless regions and specular highlights: A progressive space carving scheme using a novel photoconsistency measure. ICCV, 576-584.

[7] Seitz, S., Curless, B., Diebel, J., Scharstein, D. et Szeliski R. 2006. A comparison and evaluation of multi-view stereo reconstruction algorithms. Proc. CVPR, 519 - 528.

[8] Chang, J., Park, H., Park, I., Lee, K. et Lee, S. 2011. GPU-friendly multi-view stereo reconstruction using surfel representation and graph cuts. CVIU. 115, 5, 620-634.

[9] Deng, Y., Liu, Y., Dai, Q. et Zhang. 2012. Noisy Depth maps fusion for multi-view stereo via matrix completion. Selected Topics in Signal Processing, IEEE Journal of, 6, 5, 566-582.

[10] Guillemaut, J. et Hilton, A. 2011. Joint multi-layer segmentation and reconstruction for free-viewpoint video Applications. IJCV, 93, 1, 73-100.

[11] Lu, D., Zhu, W., Diao, C., Xiang J. et Xu, D. 2012. Continuous disparity chain estimation for multi-view stereo. Submitted to Journal of Zhejiang University Science C.

[12] Sande, K., 2004. A practical setup for voxel coloring using off-the-shelf components. University of Amsterdam, Bachelor thesis.

[13] Bakkay, M.C., Barhoumi, W. et Zagrouba, E. 2011. Mise à jour dynamique de l'image de référence pour l'optimisation du résume vidéo en ligne par multiple mosaïques. TAIMA, 135-144.

[14] Barhoumi, W., Bakkay, M.C. et Zagrouba, E. 2011. An online approach for multi-sprite generation based on camera parameters estimation. IJSIVP, Springer, 1-11.

[15] Tran, S. et Davis, L. 2006. 3D surface reconstruction using graph cuts with surface constraints. ECCV. 2, 219-231.

[16] Vogiatzis, G., Torr, P. et Cipolla, R. 2005. Multi-view stereo via volumetric graph-cuts. CVPR, 391-398.

[17] Slabaugh, G., Culbertson, W., Malzbender, T., Stevens, M. et Schafer, R. 2004. Methods for volumetric reconstruction of visual scenes. IJCV, 57, 3, 179-199.

[18] Stevens, R., Culbertson, B. et Malzbender, T. 2002. A histogram-based color consistency test for voxel coloring. ICPR, 4, 118-121.